\documentclass[a4paper,fleqn]{cas-dc}
\usepackage[numbers]{natbib}

\usepackage[T1]{fontenc}
\usepackage{graphicx}
\usepackage{booktabs}
\usepackage{amsmath,amssymb}
\usepackage{hyperref} 
\usepackage[nameinlink,noabbrev]{cleveref} 
\usepackage{array}      
\usepackage{booktabs}  
\usepackage{multirow}  
\usepackage{amsmath}   
\usepackage{booktabs}
\usepackage{multirow}
\usepackage{amsmath}
\usepackage{graphicx}  
\usepackage{svg}
\usepackage{longtable} 
\usepackage{tabularx}
\usepackage{adjustbox}
\usepackage{algorithm}
\usepackage{algpseudocode}

\svgsetup{
  inkscapeexe="D:/Inkscape/bin/inkscape.exe"
}

\title[mode = title]{A Hierarchical Error-Corrective Graph Framework for Autonomous Agents with LLM-Based Action Generation}
\author[1]{Cong Cao}[orcid=0000-0000-0000-0000]
\cormark[cor1]
\ead{magicmaster356a@gmail.com}
\author[2]{Jingyao Zhang}
\author[3]{Kun Tong}

\affiliation[1]{organization={Independent Researcher},
                city={Hangzhou},
                state={Zhejiang},
                country={China}}

\affiliation[2]{organization={Beihang University},
                city={Hangzhou},
                state={Zhejiang},
                country={China}}

\affiliation[3]{organization={Beihang University},
                city={Hangzhou},
                state={Zhejiang},
                country={China}}

\cortext[cor1]{Corresponding author: \href{mailto:magicmaster356a@gmail.com}{magicmaster356a@gmail.com}}

\date{}

\begin{document}
\shorttitle{Hierarchical Error-Corrective Graph Framework}
\shortauthors{C. Cao et al.}

\begin{abstract}
While the integration of Reinforcement Learning (RL) and Large Language Models (LLMs) has enabled agents to generate high-level action plans for complex embodied tasks, robust execution remains an open challenge. Existing approaches struggle with three critical limitations: Firstly, traditional methods typically rely on single-dimensional metrics or simple weighted scoring mechanisms, which makes it difficult to comprehensively characterize the transferability of strategies in different tasks. This limitation is particularly pronounced in dynamic or partially observable environments. Secondly, current agent feedback mechanisms often focus solely on overall task success or failure, without providing structured attribution for the causes of failure. Finally, existing Retrieval-Augmented Generation (RAG) methods have achieved some success in mitigating LLM hallucinations, but their retrieval processes primarily depend on vector similarity or token-based matching, capturing only superficial semantic proximity and failing to fully leverage the structured relationships among historical experiences, actions, and events. This limitation restricts retrieval quality, semantic alignment, and contextual consistency. To address these issues, we propose a Hierarchical Error-Corrective Graph Framework (HECG) for Autonomous Agents with LLM-Based Action Generation, which incorporates three core innovations: (1) Multi-Dimensional Transferable Strategy (MDTS): By integrating task quality, execution cost, risk,  and LLM-based semantic reasoning scores, MDTS achieves multi-dimensional alignment for the precise selection of high-quality candidate strategies, effectively reducing negative transfer risk. (2) Error Matrix Classification (EMC): Unlike simple confusion matrices, EMC provides structured attribution by categorizing errors into ten types based on severity and recoverability, offering clear and actionable guidance for subsequent error correction and replanning.(3) Causal-Context Graph Retrieval (CCGR): CCGR construct graphs from historical states, actions, and event sequences to identify subgraphs most relevant to the current context. This captures deep structural dependencies beyond flat vector similarity, allowing agents to accelerate strategy adaptation and improve execution reliability. Extensive experiments in simulated embodied environments (e.g., VirtualHome) demonstrate that HECG significantly outperforms state-of-the-art LLM planners, achieving substantial improvements in task success rates, execution efficiency, and error recovery. Ultimately, our framework bridges the gap between high-level semantic reasoning and robust low-level execution in complex, multi-step tasks. The implementation and experimental code are publicly available at:https://gitlab.com/magicmaster356a/cozy-hierplan
\end{abstract}

\maketitle

\section{Introduction}

Recent advances in autonomous robotics and multi-robot systems have enabled increasingly complex tasks in unstructured and dynamic environments. Robots are now expected to perform sequential actions, adapt to unforeseen disturbances, and collaborate efficiently with other agents. Traditional approaches often include three core aspects of autonomous/multi-robot systems: Perception, Planning, and Collaboration~\cite{zhao2024survey}.

A number of recent works have explored hybrid or learning-based solutions that partially alleviate execution uncertainty and coordination complexity. For instance, combining classical planning with reinforcement learning (RL) has shown promise in dynamic and multi-agent navigation settings. A recent study by Zhao~\cite{zhao2025voronoi} integrates Voronoi partitioning with deep RL to improve multi-robot exploration efficiency and dynamic obstacle avoidance in unknown environments. Similarly, decentralized, real-time, asynchronous probabilistic trajectory planning frameworks have demonstrated improved success rates in cluttered and dynamically changing scenarios. In industrial and applied robotics, collaborative paradigms—such as those discussed in surveys from venues like Sukhatme's research—emphasize robustness, safety, and real-world timing constraints, highlighting the need for adaptive and intelligent coordination strategies beyond idealized simulation environments~\cite{ref_dream2024}.
\begin{figure}  
  \centering
  \includegraphics[width=.9\linewidth]{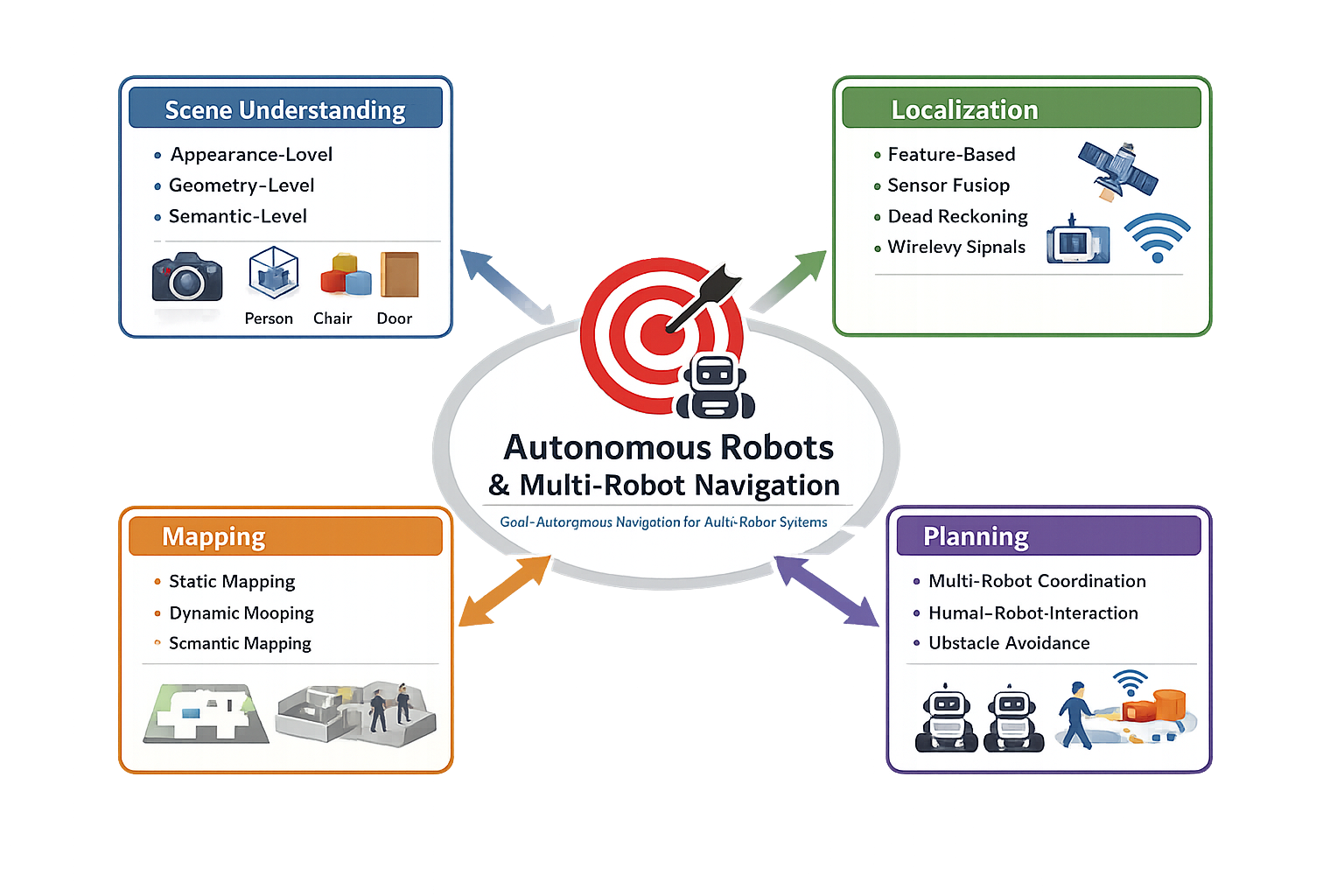} 
  \caption{A categorization of autonomous robot methods.}
  \label{fig:autonomous-robot-methods}
\end{figure}
However, while these approaches enhance adaptability and navigation robustness, they do not directly address several deeper structural limitations in transfer and error reasoning. Most existing transfer mechanisms still rely primarily on single-dimensional performance indicators (e.g., cumulative reward or success rate), which are insufficient to capture semantic compatibility and contextual alignment between source and target tasks. As a result, policy selection may remain vulnerable to negative transfer despite improved learning-based control. Likewise, execution failures are often summarized through aggregate metrics, without a structured mechanism to analyze failure type, source, and severity—limiting the system’s ability to perform systematic corrective refinement. Furthermore, retrieval or reuse of prior experience in dynamic environments typically depends on flat similarity matching, overlooking the causal and sequential dependencies embedded in historical state–action trajectories, thereby constraining generalization and long-horizon adaptability.

This paper proposes a Hierarchical Error Correction (HEC) framework that integrates large language model based action generation with structured error-driven execution. In this framework, an LLM outputs a set of optional actions (action candidates), which are executed sequentially. Execution outcomes are monitored using task-specific error metrics:
\begin{itemize}
  \item If the observed error is below a defined threshold, the system proceeds to the next step automatically.
  \item If the error exceeds the threshold, a multi-level correction mechanism is triggered.
\end{itemize}
The HECG framework operates across three hierarchical levels. At the first level, Local Correction performs fine-grained parameter adjustments to individual actions (e.g., minor end-effector repositioning). The second level, Optional Action Switching, enables strategy-level corrections by selecting alternative actions from a predefined optional set (e.g., switching from grasping to pushing when dealing with obstacles). Task Re-Planning is the third level, which provides high-level corrections by regenerating the entire action sequence while preserving historical failure information, ensuring that previously unsuccessful strategies are not repeated. A key innovation of this work is the integration of a graph-structured task representation with hierarchical error correction. Each node in the graph corresponds to an action or subgoal, encapsulating the action content, expected outcome, local threshold, and local correction rules. Edges represent task flows or optional transitions, including default paths, alternative branches, error-correction paths, and fallback mechanisms. This structure enables error-driven graph traversal, in which execution dynamically adapts to observed failures. When errors remain below predefined thresholds, the system continues along the main task path. If errors exceed these thresholds, corrective edges are activated to invoke appropriate local or strategy-level corrections. In cases of persistent or repeated failures, higher-level fallback edges are triggered, potentially leading to task re-planning or escalation to human intervention. 

By combining hierarchical correction with graph-based task representation, the proposed framework provides robust, adaptive, and interpretable decision-making for autonomous agents. The main contributions of this work are summarized as follows:

\begin{enumerate}
    \item \textbf{Multi-Dimensional Transferable Strategy (MDTS):} 
    We introduce a novel strategy selection mechanism that integrates task quality, execution cost, risk, and LLM-based semantic scores. This design enables precise strategy selection while effectively mitigating negative transfer across tasks.

    \item \textbf{Error Matrix Classification (EMC):} 
    We propose a structured error attribution mechanism that classifies task failures based on severity and recoverability. This framework provides actionable guidance for multi-level error correction and improves system robustness.

    \item \textbf{Causal-Context Graph Retrieval (CCGR):} 
    We design a graph-based retrieval method that captures both structural and causal dependencies in historical trajectories. Compared to traditional flat similarity matching, CCGR significantly enhances retrieval quality and long-horizon adaptability.
\end{enumerate}

\section{Related Work}
\subsection{LLM-based Planning for Embodied Agents}
Recent advances in large language models (LLMs) have significantly influenced task planning and decision-making for embodied agents. By leveraging rich world knowledge and strong reasoning capabilities, LLMs have been used to generate high-level action sequences for robots in manipulation, navigation, and multi-step task execution scenarios. For example, representative systems integrate LLMs with feedback loops and modular recovery mechanisms to improve robustness in manipulation and navigation tasks, formalizing task planning as a goal-conditioned Markov decision process with explicit error mitigation strategies~\cite{garrabe2024enhancing}. Other representative works such as SayCan~\cite{ahn2022grounding},  Joublin~\cite{joublin2024copal}, InteLiPlan~\cite{ly2026inteliplan} and ProgPrompt~\cite{singh2022progprompt} demonstrate how LLMs can translate natural language instructions into executable robot actions by integrating symbolic reasoning with low-level controllers. Other efforts focus on constraining the output of an LLM for safer and more reliable planning, incorporating formal logic or chain-of-thought reasoning to reduce unsafe or infeasible action proposals in service robotics domains~\cite{obi2025safeplan}. More recent research further explores structured alignment between high-level language reasoning and executable control primitives, introducing intermediate representations or action abstraction layers to ensure semantic consistency between generated plans and embodied affordances~\cite{ao2025llmbt}. Such approaches enhance controllability and execution fidelity by bridging language-based reasoning with grounded policy execution. Additionally, hierarchical LLM planners have been proposed to support multi-robot teams and event-driven replanning, demonstrating that structured decomposition and local event monitoring can enhance adaptability under execution disturbances~\cite{borate2025hierarchical}. These methods illustrate the ongoing push toward integrating symbolic reasoning and feedback awareness in LLM-based planners.

Despite these advances, many LLM planners still treat executable actions as fixed sequences with limited error-aware execution control. They often rely on coarse success/failure signals or post-hoc replanning without a systematic hierarchy of error thresholds and corrective actions during execution. In addition, these approaches typically assume reliable perception and actuation or rely on simple success/failure signals. As a result, LLM-generated plans often degrade in real-world environments due to sensor noise, execution uncertainty, or incomplete environment understanding. Several studies and surveys have reported that, without structured feedback or correction mechanisms, LLM-based planners often produce brittle plans that lack robustness under real-world disturbances. Works like Robot planning with LLMs highlight that language models alone are insufficient for responsive execution due to limited grounding in sensorimotor feedback, and must be integrated with perception and control modules to achieve reliable embodied behavior~\cite{nature2025_robotplanning,pmc2025_llm_navigation,kim2024_llm_survey}.
\subsection{Error-aware Execution and Recovery}

While LLM-based planners enhance high-level reasoning and task decomposition, robust embodied execution fundamentally depends on structured feedback and error-aware control mechanisms. Unlike the reasoning-oriented approaches discussed in Section 2.1, this line of work focuses on mitigating execution uncertainty arising from perception noise, actuation deviation, and environmental dynamics.

At the foundation of robotic robustness lies classical closed-loop control. Model-based controllers continuously incorporate sensor feedback to compensate for motion error and environmental perturbation ~\cite{siciliano2008_handbook}. These methods provide strong guarantees at the motion level, but they typically operate below the semantic task abstraction and do not explicitly reason about symbolic task failure.At the task level, structured execution frameworks such as Behavior Trees (BTs) and Finite State Machines (FSMs) introduce modular fallback and recovery policies ~\cite{colledanchise2018_bt}. By explicitly encoding failure conditions and recovery transitions, these representations improve predictability and reliability in industrial and service robotics. Their hierarchical control flow allows predefined corrective actions to be triggered under specific error states, enabling interpretable and safety-aware execution. However, their robustness largely depends on manually designed failure branches and domain-specific rules, limiting scalability to open-ended or novel tasks.

More recent research attempts to introduce adaptivity into structured planners. The SDA-PLANNER framework incorporates state dependency modeling and error-adaptive repair strategies, dynamically regenerating action subtrees based on execution feedback ~\cite{shen2025_sdaplanner}. Similarly, approaches combining LLM reasoning with semantic digital twins leverage environment-grounded simulation to iteratively refine plans and enable context-aware correction strategies, demonstrating improved robustness in embodied benchmarks such as ALFRED ~\cite{naeem2025_digitaltwins}. These efforts move beyond static fallback rules by incorporating environment-aware monitoring and partial replanning.

Learning-based methods further generalize recovery behaviors. Hierarchical reinforcement learning (HRL) decomposes complex tasks into sub-policies capable of local adaptation while maintaining global objectives~\cite{kulkarni2016hierarchical}. Imitation learning has also been explored to capture human recovery strategies in manipulation and navigation tasks. These approaches show empirical gains in adaptability compared to purely rule-based systems.

Despite these advances, several limitations persist. First, many structured frameworks rely on reactive recovery triggered by binary success/failure signals, lacking graded or multi-level error modeling. Second, learning-based recovery mechanisms often require extensive training data and remain difficult to interpret or verify. Third, although recent systems attempt to integrate LLM reasoning with execution feedback, the coupling is frequently loose: high-level plans are corrected post hoc rather than being generated with explicit awareness of execution-level uncertainty. As a result, recovery remains local and reactive, rather than being embedded within a principled hierarchy of error thresholds and corrective strategies.

Overall, existing feedback-driven execution methods significantly improve robustness at the control and subtask levels. However, a systematic integration of high-level symbolic planning with structured, multi-level error management remains underexplored. This gap motivates the development of planning frameworks that incorporate error-awareness directly into the planning-execution loop.
\subsection{Retrieval for Decision Making}

Retrieval-Augmented Generation (RAG) has emerged as an effective paradigm for mitigating hallucinations and improving contextual grounding in Large Language Models. The original RAG framework proposed by Lewis et al.~\cite{lewis2020rag} integrates dense retrieval with sequence generation, enabling models to condition outputs on external knowledge sources. Subsequent work such as REALM by Guu et al.~\cite{guu2020realm} further demonstrated the benefits of end-to-end differentiable retrieval for knowledge-intensive tasks. In embodied and decision-making settings, retrieval mechanisms have been increasingly adopted to provide contextual grounding for high-level planning and reasoning.

However, most existing RAG-based approaches rely primarily on vector similarity search, typically implemented using dense embeddings and nearest-neighbor retrieval. While effective for semantic matching, such methods often overlook structural, causal, and temporal dependencies embedded in embodied trajectories. In sequential decision-making tasks, particularly under partially observed or dynamic environments, the relationships among states, actions, and events are inherently compositional and structured. Purely embedding-based retrieval may retrieve semantically similar but structurally incompatible trajectories, leading to suboptimal policy adaptation or inconsistent planning.To address these limitations, recent research has explored structured memory representations. Memory-augmented agents such as DeepMind’s Gato~\cite{reed2022gato} and generative agents proposed by Park et al.~\cite{park2023generative} demonstrate how structured memory and event abstraction can improve long-horizon coherence and decision consistency. In robotics and embodied AI, approaches leveraging scene graphs, task graphs, and relational world models encode interactions among objects and actions as graph structures, enabling reasoning over affordances and causal dependencies rather than isolated tokens or embeddings~\cite{johnson2015scenegraph}.

Causal-Context Graph Retrieval (CCGR) further extend this idea by organizing historical trajectories into nodes (states, actions, subgoals) and edges (temporal transitions, causal effects, or semantic relations). Compared to flat vector databases, graph-structured memories preserve higher-order dependencies, enabling subgraph matching and structural similarity search~\cite{battaglia2018graphnet,jiang2022graphrl}. This allows agents to retrieve not only semantically related experiences but also structurally aligned execution patterns, which is particularly beneficial for multi-step task adaptation and transfer learning~\cite{pritzel2017neuralepisodic,blundell2016modelfree}.Despite these advances, integration between retrieval and transfer evaluation remains limited. Existing methods often treat retrieval as a preprocessing step for generation, without jointly considering policy transferability, error profiles, or execution uncertainty. Moreover, vector-based retrieval alone cannot capture multi-dimensional transfer signals such as reward trends, confidence metrics, or LLM-derived semantic consistency scores ~\cite{yao2022react}. In light of these limitations, our framework introduces Graph-Based Retrieval (Graph Retrieve) to structure historical states, actions, and event sequences into a graph memory. By performing subgraph-level relevance matching conditioned on current task context, the system captures structural dependencies beyond embedding similarity. Combined with our multi-dimensional transfer evaluation (Q, C, R, LLM-Score) and structured error matrix classification, this approach enables more reliable experience reuse, reduces negative transfer, and enhances execution robustness in complex embodied environments.

\section{Method}
\subsection{Overview}

Large Language Models (LLMs) have shown strong capabilities in high-level planning and sequential reasoning, enabling embodied agents to translate natural language instructions into action sequences. However, a key challenge remains: the mismatch between generated plans and environment dynamics often leads to execution failures, especially in long-horizon tasks.

This challenge arises from the \emph{plan–environment alignment gap}, where the LLM’s implicit world model diverges from the grounded simulator state. Such misalignment can manifest as incorrect assumptions about object states, missing environmental preconditions, or stochastic execution outcomes. In long-horizon settings, these deviations accumulate over time, causing cascading failures and reducing overall task reliability.

To address this issue, we focus on \emph{robust execution of LLM-generated plans} rather than improving plan generation itself. Given a natural language instruction $T$ and an LLM-generated action sequence
\begin{equation}
\pi = (a_1, a_2, \dots, a_n),
\end{equation}
our goal is to execute $\pi$ reliably in stochastic and partially observable environments, minimizing failure propagation and unnecessary global replanning.

We propose the \emph{Hierarchical Error-Corrective Control Graph (HECG)} framework, which reformulates plan execution as a \emph{graph-based, feedback-driven process}. Specifically, we represent the plan as a directed graph
\begin{equation}
G = (V, E),
\end{equation}
where nodes correspond to executable actions or subgoals, and edges encode transitions conditioned on execution outcomes and error types. This formulation enables dynamic adaptation during execution, rather than rigidly following a predefined sequence.

HECG integrates three key mechanisms. First, an \emph{Error Matrix Classification (EMC)} module categorizes execution failures based on severity and recoverability, enabling structured error attribution. Second, a \emph{Causal-Context Graph Retrieval (CCGR)} module retrieves relevant subgraphs from past trajectories, allowing the agent to reuse structured recovery strategies. Third, a \emph{Multi-Dimensional Transfer Strategy (MDTS)} selects candidate actions or policies by jointly considering task quality, execution cost, risk, and semantic alignment.These components are unified through a \emph{hierarchical transition policy} that organizes recovery into three levels: local action correction, alternative action selection, and task-level replanning. As a result, minor deviations are handled locally, while more severe inconsistencies trigger higher-level adaptations.

By combining structured error classification, graph-based retrieval, and hierarchical control, HECG transforms brittle sequential execution into a \emph{robust, adaptive process}, improving long-horizon stability and reducing reliance on costly global replanning.

\subsection{Graph-Based Retrieval}
We store experience as a graph $G=(V,E)$ where nodes represent states, observations, actions, and outcomes, and edges encode temporal and causal relations (``action causes transition'', ``state enables action'', ``failure leads to recovery''). Given a current context, retrieval aims to identify a subgraph that matches both semantic intent (goal and constraints) and structural similarity (relevant transition patterns). The retrieved subgraph provides (i) candidate actions, (ii) known failure modes, and (iii) recovery patterns to condition the LLM planner and the execution controller.

Each node $v_i \in V$ in the HECG represents an executable action or an intermediate subtask, serving as the fundamental unit for modeling both nominal execution and error-aware recovery. Formally, a node is defined as a tuple: 
\begin{equation}
v_i = \langle t_i, a_i, \hat{o}_i, \epsilon_i, C_i, n_i \rangle
\end{equation}
where each component explicitly captures different aspects of action execution under uncertainty. Specifically, $t_i$ encodes task-level semantic information, including the subtask name and the set of relevant environment objects, grounding high-level task intent into a concrete execution context. The action $a_i$ denotes the executable primitive or skill (e.g., \emph{grasp}, \emph{move}, \emph{push}), which can be directly issued to the low-level controller or motion planner. The expected outcome $\hat{o}_i$ specifies the desired post-condition of the action, such as an object being grasped or a target location being reached, and serves as the reference for monitoring execution success. 

The local error threshold $\epsilon_i$ defines acceptable deviation bounds between the observed execution outcome and the expected outcome, enabling the system to distinguish between minor execution noise and significant failures. When deviations exceed this threshold, a set of predefined local correction rules $C_i$ can be triggered to perform lightweight recovery actions, such as reattempting a grasp or adjusting the approach trajectory, without immediately invoking global replanning. Finally, $n_i$ specifies the subsequent connected node(s), allowing the graph to represent both linear progressions and branching execution paths. 

Transitions between nodes are encoded as directed edges $e_{ij} \in E$, which are defined as:
\begin{equation}
E \subseteq V \times K \times V, \quad e_{ij}^k = (v_i, k, v_j), \quad k \in K
\end{equation}
Each edge captures not only the source and destination nodes but also the semantic type of the transition, explicitly modeling different execution and recovery behaviors. The edge type set is defined as:
\begin{equation}
K = \{\text{main}, \text{opt}, \text{corr}, \text{fb}\}
\end{equation}
which includes: 
\begin{itemize}
    \item \textbf{Main execution edges}: define the nominal task flow generated by the LLM.
    \item \textbf{Optional edges}: connect alternative actions or skills that achieve the same subgoal, providing redundancy under execution uncertainty.
    \item \textbf{Correction edges}: activated when local errors exceed the predefined threshold, enabling targeted, node-level recovery.
    \item \textbf{Fallback edges}: allow the system to escalate from local correction to higher-level recovery strategies, such as switching subtask order or invoking task-level replanning.
\end{itemize}

By explicitly separating nominal execution, local correction, and high-level fallback mechanisms within the graph structure, HECG enables structured reasoning over failure modes and recovery strategies. This representation allows the agent to adapt its behavior dynamically based on real-time execution feedback, improving robustness while maintaining interpretability and reducing unnecessary global replanning.

\subsection{Multi-Dimensional Transfer Strategy}
\subsubsection{Transition Policy}

Given the HECG representation, where nodes encode executable actions or subtasks and edges encode nominal, optional, corrective, and fallback transitions, the remaining challenge is to determine which outgoing edge should be activated at each execution step under partial observability and execution uncertainty. Rather than relying on a fixed priority ordering or deterministic rules, we associate each candidate transition $e_{ij}^k$ with a probabilistic transition policy that dynamically evaluates its suitability based on both structured signals and high-level semantic reasoning.

Specifically, at runtime, the agent maintains a belief state $b_t$ capturing task progress and abstract execution context, as well as a low-level observation $o_t$ reflecting the current environment and robot state. For each outgoing edge from the current node $v_i$, the agent computes a transition probability that integrates task value, execution cost, risk awareness, and LLM-based semantic feasibility. This policy formulation enables the HECG graph to function not merely as a static control structure, but as an adaptive decision graph that responds continuously to execution feedback. Formally, the probability of selecting a transition from node $v_i$ to node $v_j$ with semantic type $k$ is defined as:
\begin{equation}
\begin{aligned}
\pi_{ij}^k(b_t, o_t) =
\text{Softmax}\Big(
&\alpha Q_{ij}^k(b_t)
- \beta C_{ij}(o_t)  \\
&- \gamma R_{ij}(o_t)
+ \lambda \Phi_{ij}^{\text{LLM}}(b_t, o_t)
\Big)
\end{aligned}
\end{equation}
This formulation decomposes transition selection into complementary components. The value term $Q_{ij}^k(b_t)$ captures long-horizon task utility and progress within the HECG graph, analogous to an MDP-style action-value function defined over abstract belief states. The execution cost $C_{ij}(o_t)$ penalizes transitions that are inefficient in terms of time, energy, or motion complexity, while the risk term $R_{ij}(o_t)$ estimates failure likelihood or safety hazards based on current observations. Crucially, the LLM-based score $\Phi_{ij}^{\text{LLM}}(b_t, o_t)$ injects semantic and commonsense reasoning into the decision process, allowing the agent to prefer transitions that are logically consistent with task intent, object affordances, and causal constraints that may not be explicitly encoded in low-level models. The coefficients $\alpha, \beta, \gamma, \lambda$ control the relative influence of these factors.

Each selected transition is grounded in a low-level skill $\kappa_{ij} \in \mathcal{K}_{\text{skill}}$, whose execution induces a stochastic observation transition, thereby closing the loop between symbolic decision-making and continuous control:
\begin{equation}
p(o_{t+1} \mid o_t, \kappa_{ij})
\end{equation}
\paragraph{Example: Transition Selection with LLM Participation.} Consider a node $v_i$ corresponding to the subtask ``pick up a mug from the table'', with three outgoing transitions:

\begin{itemize}
    \item Main edge $e_{ij}^{\text{main}}$: grasp mug directly.
    \item Optional edge $e_{ik}^{\text{opt}}$: reposition gripper and then grasp.
    \item Fallback edge $e_{il}^{\text{fb}}$: clear surrounding objects and retry.
\end{itemize}

Suppose the robot observes that the mug is partially occluded. The direct grasp transition has high task value $Q$ but also elevated risk $R$. The optional transition incurs slightly higher cost $C$ but lower risk. Meanwhile, the LLM assigns a high semantic feasibility score $\Phi^{\text{LLM}}$ to the optional transition, reasoning that ``adjusting the gripper before grasping is appropriate when the object is occluded.'' The fallback transition receives a lower LLM score, as clearing the table is semantically unnecessary at this stage. After combining all terms through the Softmax policy, the optional edge achieves the highest probability and is selected. If this transition later fails repeatedly, the accumulated risk increases and probability mass naturally shifts toward the fallback edge, enabling escalation without explicit rule-based switching. This example illustrates how HECG integrates structured decision signals and LLM reasoning to achieve adaptive and interpretable transition selection.
\begin{figure*}
  \centering
  \includegraphics[width=.7\textwidth]{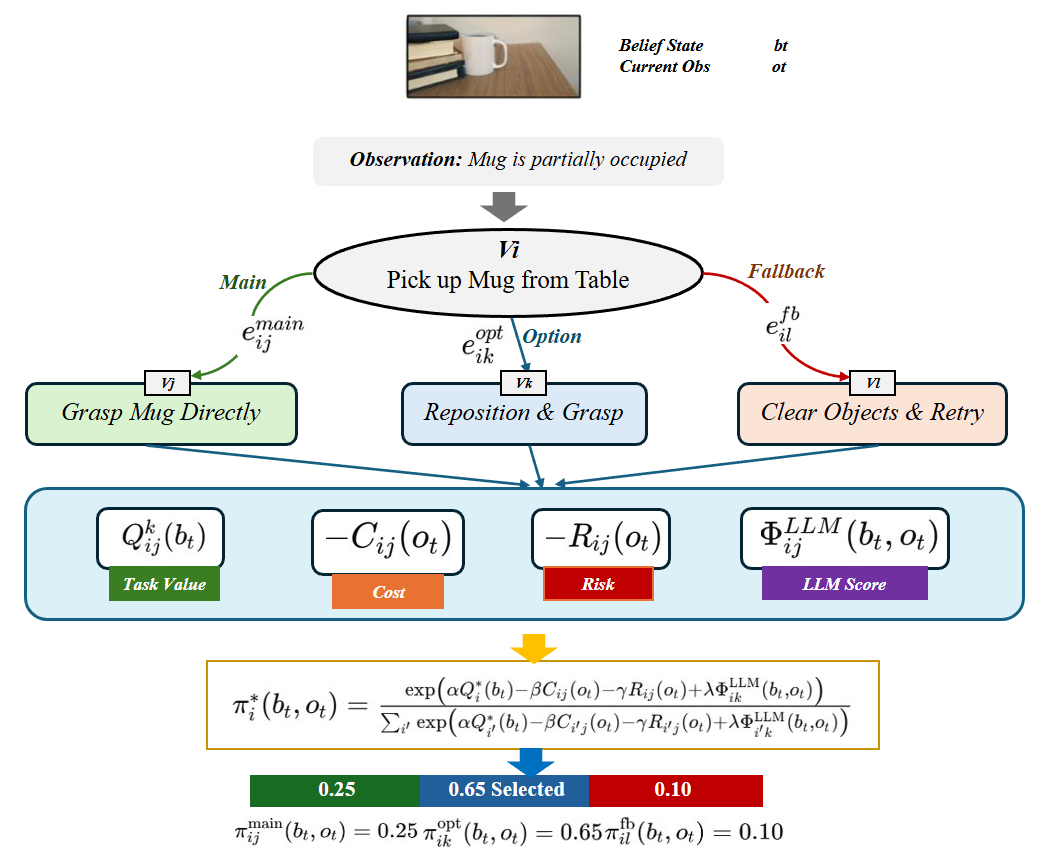}
  \caption{Structure of HECG Transition Policy.The agent selects among alternative action strategies by integrating task value, cost, risk, and LLM-based scores, and determines the final action via a softmax policy under partial observability.}
  \label{fig:transition-policy-methods}

\end{figure*}

\subsubsection{Control-Theoretic View of Error-Triggered Transitions}

To explicitly model execution deviations and enable robust error handling, we define the local execution error for a node $v_i$ as
\begin{equation}
e_i = \delta(o_t, \hat{o}_i)
\end{equation}
where $o_t$ is the current observation and $\hat{o}_i$ is the expected outcome of the action associated with $v_i$. This error metric quantifies the discrepancy between actual and desired outcomes, providing a principled basis for activating different transitions in the HECG. Using this error signal, transitions are triggered according to a state-dependent, threshold-based policy:
\begin{equation}
\pi_{ij}^k =
\begin{cases}
1, & k = \text{main}, \ e_i \le \epsilon_i\\[2mm]
2, & k = \text{corr}, \ \epsilon_i < e_i \le \epsilon_i^{\max}\\[1mm]
3, & k = \text{fb}, \ e_i > \epsilon_i^{\max}\\
0, & \text{otherwise}
\end{cases}
\end{equation}
where $\epsilon_i$ and $\epsilon_i^{\max}$ denote the local and maximum tolerable error thresholds, respectively. Intuitively, main edges are taken when execution is within acceptable bounds, correction edges are activated when the deviation exceeds local tolerance but remains recoverable, and fallback edges are triggered when errors surpass the maximum threshold, prompting higher-level replanning or recovery. This formulation naturally yields a hybrid switching system, where the HECG transitions are governed by state-dependent guards based on real-time execution feedback.

From a formal perspective, the HECG can be interpreted as a guarded finite-state automaton
\begin{equation}
A = (V, E, \Sigma, G)
\end{equation}
where $V$ and $E$ correspond to the nodes and edges of the graph, $\Sigma$ is the observation alphabet, and $G(e_{ij})$ denotes guard conditions defined over belief and error states. Each guard encodes the threshold logic described above, enabling the agent to switch between nominal, corrective, and fallback behaviors in a structured and interpretable manner. This automaton view highlights the HECG's dual nature as both a planning graph and a control system, bridging symbolic task reasoning with control-theoretic robustness.

\subsection{Error Matrix Classification}
\subsubsection{Three-Level Error Correction}

We define an error matrix that indexes failures by \textit{type} (e.g., perception, planning, control), \textit{source} (sensor noise, occlusion, actuation slip, constraint violation), and \textit{severity} (recoverable locally vs.\ requiring replanning). During execution, observed error signals are mapped into the matrix to decide the correction level:
\begin{itemize}
  \item \textbf{L1 Local correction:} small continuous adjustments (re-grasp offsets, trajectory refinement).
  \item \textbf{L2 Option switching:} choose an alternative action or policy from a candidate set.
  \item \textbf{L3 Task replanning:} regenerate the plan with constraints derived from failure history.
  \item \textbf{L4 Human-in-the-loop:} safety or repeated failure triggers escalation.
\end{itemize}

The first level of correction focuses on local, low-cost recovery strategies that do not alter the global task structure. When the error slightly exceeds the predefined threshold, the system applies a set of local correction rules $C_i$ associated with the current action node. These rules typically include continuous or parameter-level adjustments, such as fine-tuning the end-effector pose, reattempting a grasp with modified force or orientation parameters, or refining motion trajectories to compensate for minor disturbances. Since these corrections are computationally inexpensive and fast to execute, they enable rapid recovery from small, recoverable deviations while maintaining execution efficiency and stability.

If local corrections repeatedly fail or if the error magnitude exceeds a higher escalation threshold, the system transitions to optional action switching. At this level, the agent selects an alternative action node connected via optional edges in the Hierarchical Execution Control Graph (HECG). These optional actions represent discrete strategy variations that remain compatible with the overall task objective. For instance, if a direct grasp action fails persistently, the agent may switch to pushing the object into a more favorable configuration, repositioning the robot base, or approaching the object from a different direction. Optional actions can be generated dynamically by a Large Language Model (LLM) during planning or predefined based on domain expertise. This level enables structured strategy adaptation without triggering a full task replanning process.

When both local action correction and optional action switching fail to resolve the execution error, the system escalates to task-level replanning. At this stage, the complete execution context—including accumulated failure history, rejected action nodes, and updated environmental constraints—is fed back into the LLM to synthesize a revised task plan. Crucially, actions that have previously failed are explicitly annotated to prevent their repetition in the newly generated plan. The updated plan is then re-encoded into a revised HECG, allowing execution to resume with improved robustness and informed decision-making. In extreme cases involving persistent failure or safety-critical conditions, the system may further escalate to a human-in-the-loop intervention node, ensuring safe and controlled recovery.

\subsubsection{Error-Driven Graph Traversal Algorithm}

Execution follows an error-driven traversal over the Hierarchical Error Correction Graph (HECG). The process begins by initializing the traversal at the root node, after which the action associated with the current node is executed in the environment. The system then evaluates the execution outcome by computing an error metric and comparing it against a predefined threshold. Based on the magnitude and type of the observed error, the hierarchical correction policy is invoked to select the most appropriate outgoing edge, determining whether to proceed, correct, or replan at a higher level of abstraction. The execution history is continuously updated to preserve contextual information for subsequent decisions, and this cycle repeats until the task is successfully completed or a termination condition is reached. By structuring execution as an adaptive traversal rather than a fixed sequence of actions, this mechanism enables the control flow to respond robustly to real-world uncertainties and execution failures.

\subsubsection{Error Classification}

\begin{algorithm}[t]
\caption{HECG Error Handling Mechanism}
\label{alg:error-handling}
\small
\begin{algorithmic}[1]

\State Execute action $a_t$
\State Observe result $o_t$

\If{ActionNameMismatch($a_t$)}
    \State $a_t' \gets$ CorrectActionName($a_t$)
    \State \Return Transition($a_t'$)

\ElsIf{ScriptParsingError($a_t$)}
    \State \Return Failure

\ElsIf{ActionExecutionError($o_t$)}
    \If{Recoverable($o_t$)}
        \State $a_t' \gets$ AdjustTargetOrAction($a_t$)
        \State \Return Transition($a_t'$)
    \Else
        \State \Return Failure
    \EndIf

\ElsIf{SensorFailure($o_t$)}
    \State ReinitializeSensors()
    \State \Return Transition($a_t$)

\ElsIf{CollisionDetected($o_t$)}
    \State DropAction($a_t$)
    \State \Return Transition(NextAction())

\ElsIf{Timeout($o_t$)}
    \State Retry($a_t$)
    \State \Return Transition($a_t$)

\ElsIf{HardwareFault($o_t$)}
    \State EmergencyStop()
    \State NotifyHuman()
    \State \Return Failure

\ElsIf{PerceptionMismatch($o_t$)}
    \State $a_t' \gets$ RecalibrateAndRetry($a_t$)
    \State \Return Transition($a_t'$)

\ElsIf{PositioningError($o_t$)}
    \State $a_t' \gets$ RepositionAgent($a_t$)
    \State \Return Transition($a_t'$)

\Else
    \State \Return Success
\EndIf

\end{algorithmic}
\end{algorithm}

To systematically handle execution failures, we categorize errors that may arise during task execution. Table~\ref{tab:error-classification} summarizes typical error types, severity levels, associated actions, and recoverability. Additional error categories are included beyond the initial four examples to cover more realistic scenarios in embodied execution.

This extended classification helps the HECG system distinguish between recoverable, partially recoverable, and unrecoverable failures, guiding the appropriate correction strategy—from local action adjustment to optional action switching, or task-level replanning.

\section{Experiments}
\subsection{Tasks and Environments}
All experiments are conducted in a simulated environment designed to evaluate hierarchical task planning and execution under realistic uncertainty. The widely-used embodied AI and task planning dataset \textit{VirtualHome} provides programmatic household scenarios with annotated action sequences and includes diverse task types, ranging from simple single-object manipulations to complex multi-room interaction sequences. This makes it a comprehensive testbed for evaluating our proposed Hierarchical Error-Correcting Graph (HECG) framework.

In our experiments, we consider a set of representative tasks, including \textsc{ReadBook}, \textsc{PutDishwasher}, \textsc{PrepareFood}, and \textsc{PutFridge}, across different scenes such as \textsc{Bedroom}, \textsc{LivingRoom}, \textsc{Kitchen}, and \textsc{Bathroom}. These tasks require multi-step reasoning, sequential dependencies, and interaction with dynamic objects. Each task is represented as a hierarchy of nodes, where each node corresponds to either a primitive action or a higher-level subtask. During execution, observations are generated at each step and compared against expected outcomes to detect potential errors, following the formulation described in Section~3.

We implement the proposed Hierarchical Error-Correcting Graph (HECG) agent using a modular architecture that integrates three core components: planning, verification, and replanning.For the LLM-based baseline, we employ a large language model (LLM) to directly generate a sequence of actions given the task description, without incorporating explicit hierarchical error correction or transition policies.

We compare the HECG framework against the following baselines:

\begin{itemize}
    \item \textbf{LLM Planner (Flat):} A conventional LLM-based planner that generates the entire action sequence in a single pass, without hierarchical correction or replanning mechanisms.
    
    \item \textbf{HECG w/o Transition:} A variant of our model that includes hierarchical error correction but removes the learned transition policy, allowing us to measure the impact of explicit state transitions between subtasks.
    
    \item \textbf{HECG Full:} The complete HECG agent with both hierarchical correction and transition policies enabled.
\end{itemize}

\subsection{Baselines}
\begin{figure*}[htbp]
    \centering
    \includegraphics[width=0.8\textwidth]{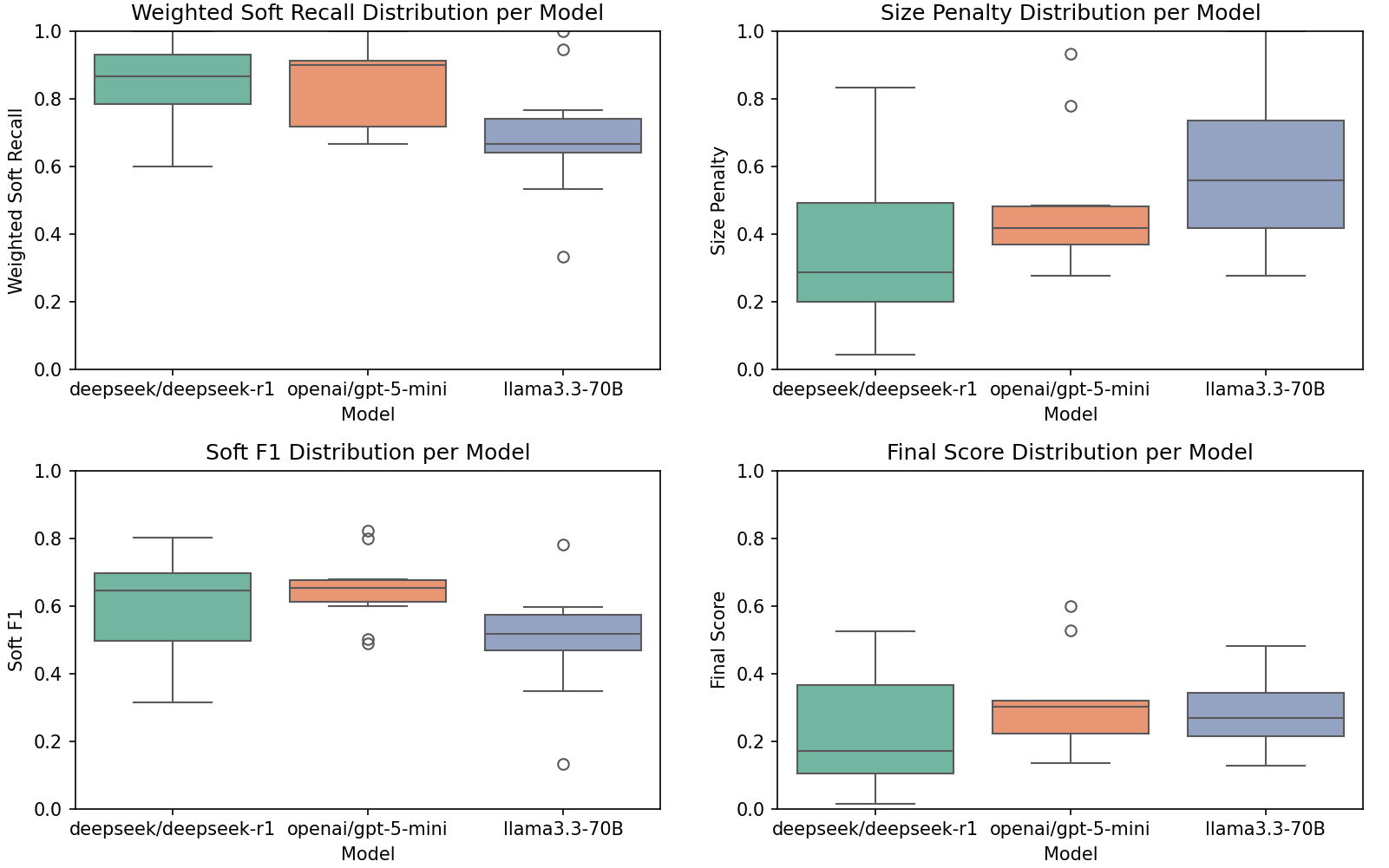}
    \caption{LLM Goal Compliance Evaluation Results.}
    \label{fig:goal-compliance-evaluation-results}
\end{figure*}
All experiments are conducted in a simulated environment designed to evaluate hierarchical task planning and execution under realistic uncertainty. The widely-used embodied AI and task planning dataset \textit{VirtualHome} provides programmatic household scenarios with annotated action sequences and includes diverse task types, ranging from simple single-object manipulations to complex multi-room interaction sequences. This makes it a comprehensive testbed for evaluating our proposed Hierarchical Error-Correcting Graph (HECG) framework.

In our experiments, we consider a set of representative tasks, including \textsc{ReadBook}, \textsc{PutDishwasher}, \textsc{PrepareFood}, and \textsc{PutFridge}, across different scenes such as \textsc{Bedroom}, \textsc{LivingRoom}, \textsc{Kitchen}, and \textsc{Bathroom}. These tasks require multi-step reasoning, sequential dependencies, and interaction with dynamic objects. Each task is represented as a hierarchy of nodes, where each node corresponds to either a primitive action or a higher-level subtask. During execution, observations are generated at each step and compared against expected outcomes to detect potential errors, following the formulation described in Section~3.

We implement the proposed Hierarchical Error-Correcting Graph (HECG) agent using a modular architecture that integrates three core components: planning, verification, and replanning.For the LLM-based baseline, we employ a large language model (LLM) to directly generate a sequence of actions given the task description, without incorporating explicit hierarchical error correction or transition policies.

We compare the HECG framework against the following baselines:

\begin{itemize}
    \item \textbf{LLM Planner (Flat):} A conventional LLM-based planner that generates the entire action sequence in a single pass, without hierarchical correction or replanning mechanisms.
    
    \item \textbf{HECG w/o Transition:} A variant of our model that includes hierarchical error correction but removes the learned transition policy, allowing us to measure the impact of explicit state transitions between subtasks.
    
    \item \textbf{HECG Full:} The complete HECG agent with both hierarchical correction and transition policies enabled.
\end{itemize}

\subsection{Overall Performance Comparison (Goal Compliance \& Task Plan)}

\paragraph{Goal Compliance Evaluation.}

We first evaluate \emph{goal compliance}, which measures the extent to which the final world state satisfies the intended objectives of a task. Unlike task success rate, which requires all actions to be executed correctly, goal compliance focuses on whether the desired end state is achieved, even if intermediate steps are suboptimal.Formally, for a task $T$ with a set of goal conditions 

\begin{equation}
G = \{g_1, g_2, \dots, g_n\},
\end{equation}
goal compliance is defined as:
\begin{equation}
\text{Goal Compliance} =
\frac{
|\{ g \in G \mid g \text{ satisfied in final state} \}|
}{|G|}
\end{equation}
where $|G|$ denotes the total number of goal conditions, and the numerator counts the number of satisfied conditions after execution.

We evaluated three models (GPT-5 Mini, DeepSeek-R1, and LLaMA3.3-70B) on multi-room household task execution scenarios. The tasks involve cross-room object interactions, including read book, put dishwasher, prepare food, put fridge, and setup table. Evaluation metrics include weighted soft recall, soft precision, soft F1, size penalty, and a final composite score reflecting overall execution quality adjusted for sequence length.


Across all evaluated tasks, GPT-5 Mini demonstrates the strongest overall performance, achieving the highest average final score (0.315), compared with LLaMA3.3-70B (0.278) and DeepSeek-R1 (0.230). GPT-5 Mini maintains a strong balance between recall and precision, with an average weighted soft recall of 0.842 and soft precision of 0.534, resulting in the highest average soft F1 (0.651) among the evaluated models.
\begin{figure*}[htbp]
    \centering
    \includegraphics[width=0.7\textwidth]{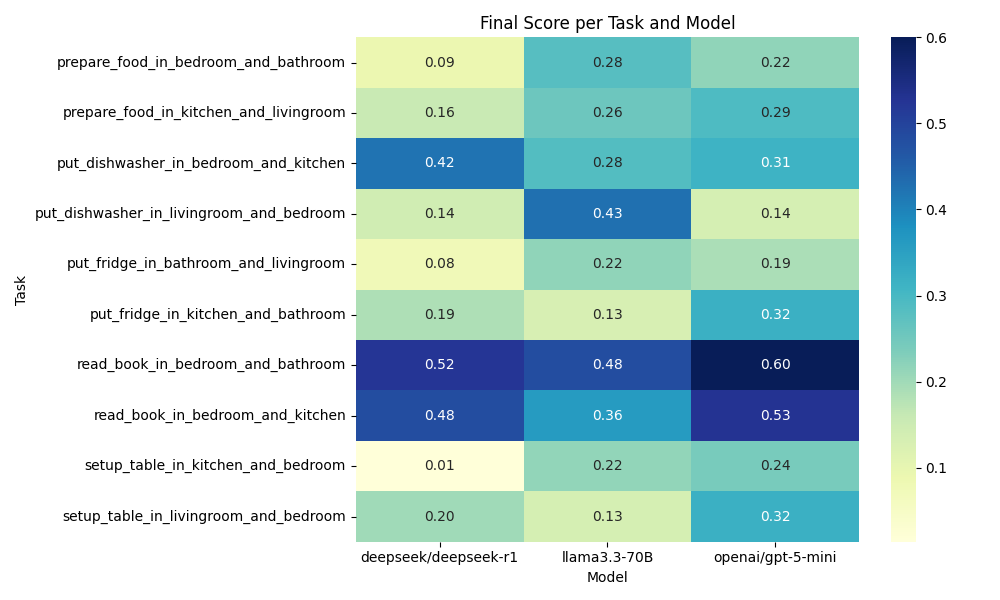}
    \caption{Heatmap of Goal Compliance.}
    \label{fig:goal-compliance-heatmap}
\end{figure*}
In particular, GPT-5 Mini performs best on put fridge in kitchen and bathroom, achieving an F1 score of 0.823 and a final score of 0.320, and also on setup table in livingroom and bedroom, where it reaches perfect recall (1.0) with an F1 score of 0.800 and a final score of 0.320. However, tasks such as put dishwasher in livingroom and bedroom yield lower scores (final score 0.135), likely due to increased spatial complexity and multi-step dependencies, which reduce precision and incur stronger sequence length penalties.

DeepSeek-R1 achieves the highest average recall (0.852) among the models and performs strongly on tasks such as put dishwasher in bedroom and kitchen (final score 0.422). However, it frequently generates longer action sequences, resulting in larger size penalties, which significantly reduce its overall composite score.

LLaMA3.3-70B demonstrates relatively balanced behavior but tends to produce longer plans with lower precision (average 0.400), which limits its final performance. While it performs competitively on tasks such as put dishwasher in livingroom and bedroom (final score 0.427), it also exhibits unstable behavior on some tasks (e.g., setup table in livingroom and bedroom, final score 0.133).

Overall, GPT-5 Mini achieves the best balance between recall, precision, and sequence efficiency, indicating stronger grounding and structured sequence planning capability. DeepSeek-R1 tends to prioritize recall at the expense of efficiency, while LLaMA3.3-70B exhibits moderate recall but reduced precision and longer action sequences.

\paragraph{Task Success and Action-Level Evaluation.}

We further conduct a comparative evaluation of basic household task execution across paired scenes (e.g., bedroom\_and\_kitchen, kitchen\_and\_bathroom). The evaluation metrics include Task Success Rate (\%), Average Action Accuracy (\%), and Total Steps.

\subparagraph{Success Rate (SR).}
Success rate measures the proportion of episodes that achieve the task goal. We distinguish between two types:
\begin{itemize}
    \item \textbf{Original Success Rate}: The success rate of executing the initially generated plan without any replanning.
    \item \textbf{Replan Success Rate}: The success rate achieved after incorporating replanning and error recovery mechanisms.
\end{itemize}
The final success rate for a task is defined as:
\begin{equation}
\mathrm{SR}_{\text{final}} = \frac{N_{\text{successful episodes}}}{N_{\text{total episodes}}}
\end{equation}

\subparagraph{Action Accuracy (AA)}
Action Accuracy measures the correctness of individual actions executed by the agent in a task episode. It is defined as the ratio of correctly executed actions to the total number of actions taken:

\begin{equation}
AA = \frac{N_{\text{correct actions}}}{N_{\text{executed actions}}}
\end{equation}

where:
\begin{itemize}
    \item $N_{\text{correct actions}}$ is the number of actions that match the reference or ground-truth action sequence.
    \item $N_{\text{executed actions}}$ is the total number of actions executed by the agent in the episode.
\end{itemize}

A value of 1 indicates that all actions in the episode were executed correctly, while lower values reflect incorrect or mismatched actions. Action Accuracy provides a fine-grained, step-level measure of procedural correctness, complementing task-level success metrics.

\subparagraph{Efficiency Score.}
Efficiency quantifies how optimally the plan achieves the task goal, measured by the number of executed actions relative to the optimal plan length:
\begin{equation}
\mathrm{Efficiency} = \frac{N_{\text{optimal steps}}}{N_{\text{executed steps}}}
\end{equation}
A value of 1 indicates perfect efficiency matching the optimal plan, while lower values indicate additional steps due to errors or inefficient recovery actions.

\subparagraph{Stability Analysis.}
We assess the consistency of model performance using the coefficient of variation (CV) of success rates across episodes:
\begin{equation}
\mathrm{CV} = \frac{\sigma_{\text{SR}}}{\mu_{\text{SR}}}
\end{equation}
where $\sigma_{\text{SR}}$ is the standard deviation and $\mu_{\text{SR}}$ is the mean of success rates across episodes. Lower CV values indicate more stable and reliable performance.

\subparagraph{Task Complexity Assessment.}
To analyze the relationship between task difficulty and model performance, we compute an overall complexity score for each task based on:
\begin{itemize}
    \item Number of required actions
    \item Environmental constraints and object interactions
    \item Sequential dependencies between subgoals
    \item Potential error-prone steps
\end{itemize}
This complexity measure enables analysis of how different models scale with task difficulty.Task complexity significantly affects step count. For instance, \emph{Preparefood} tasks require 15--19 steps, reflecting increased dependency chains. LLaMA3.3-70B generally uses fewer steps than DeepSeek-R1 for comparable tasks, suggesting relatively more efficient action sequencing.

\begin{table}[t]
\centering

\caption{Original Task Performance Metrics Across Models.
(B1: Bedroom 1, B2: Bathroom 2, K: Kitchen, L: Living Room)}
\label{tab:success_rate}
\small
\setlength{\tabcolsep}{4pt} 
\renewcommand{\arraystretch}{0.95} 
\begin{adjustbox}{margin=0cm 0cm 0cm 0cm}
\begin{tabular}{p{1.7cm} p{1.9cm} p{1.4cm} p{1.7cm} p{0.7cm}}
\hline
\textbf{Task (Scene)} & \textbf{Model} & \textbf{Original Success Rate (\%)} & \textbf{Avg. Action Accuracy (\%)} & \textbf{Total Steps} \\
\hline

\multirow{3}{*}{\makecell[l]{Readbook \\ (B1\&K)}}
& Llama3.3-70B & 0.797 & 0.833 & 7 \\
& Deepseek-R1 & 0.854 & 0.725 & 8 \\
& GPT-5-mini & 0.767 & 0.643 & 5 \\
\hline

\multirow{3}{*}{\makecell[l]{Readbook \\ (B1\&B2)}}
& Llama3.3-70B & 0.765 & 0.677 & 6 \\
& Deepseek-R1 & 0.831 & 0.933 & 8 \\
& GPT-5-mini & 0.722 & 0.744 & 7 \\
\hline

\multirow{3}{*}{\makecell[l]{Putdishwasher \\ (B1\&K)}}
& Llama3.3-70B & 0.711 & 0.866 & 9 \\
& Deepseek-R1 & 0.792 & 0.867 & 13 \\
& GPT-5-mini & 0.645 & 0.755 & 10 \\
\hline

\multirow{3}{*}{\makecell[l]{Putdishwasher \\ (L\&B1)}}
& Llama3.3-70B & 0.761 & 0.833 & 12 \\
& Deepseek-R1 & 0.828 & 0.713 & 11 \\
& GPT-5-mini & 0.715 & 0.856 & 10 \\
\hline

\multirow{3}{*}{\makecell[l]{Preparefood \\ (K\&L)}}
& Llama3.3-70B & 0.734 & 0.833 & 19 \\
& Deepseek-R1 & 0.808 & 0.865 & 15 \\
& GPT-5-mini & 0.677 & 0.955 & 17 \\
\hline

\multirow{3}{*}{\makecell[l]{Preparefood \\ (B1\&B2)}}
& Llama3.3-70B & 0.734 & 0.833 & 18 \\
& Deepseek-R1 & 0.808 & 0.725 & 16 \\
& GPT-5-mini & 0.677 & 0.968 & 17 \\
\hline

\multirow{3}{*}{\makecell[l]{Putfridge \\ (B2\&L)}}
& Llama3.3-70B & 0.749 & 0.885 & 19 \\
& Deepseek-R1 & 0.819 & 0.756 & 15 \\
& GPT-5-mini & 0.699 & 0.643 & 15 \\
\hline

\multirow{3}{*}{\makecell[l]{Putfridge \\ (K\&B2)}}
& Llama3.3-70B & 0.730 & 0.856 & 18 \\
& Deepseek-R1 & 0.806 & 0.725 & 14 \\
& GPT-5-mini & 0.671 & 0.855 & 17 \\
\hline

\multirow{3}{*}{\makecell[l]{Setuptable \\ (K\&B1)}}
& Llama3.3-70B & 0.748 & 0.844 & 17 \\
& Deepseek-R1 & 0.819 & 0.885 & 15 \\
& GPT-5-mini & 0.698 & 0.889 & 16 \\
\hline

\multirow{3}{*}{\makecell[l]{Setuptable \\ (L\&B1)}}
& Llama3.3-70B & 0.760 & 0.755 & 14 \\
& Deepseek-R1 & 0.827 & 0.725 & 16 \\
& GPT-5-mini & 0.714 & 0.855 & 17 \\
\hline

\end{tabular}
\end{adjustbox}
\end{table}

\subparagraph{Improvement Metrics.}
We quantify the benefit of replanning through improvement in success rate:
\begin{equation}
\mathrm{Improvement} = \mathrm{SR}_{\text{replan}} - \mathrm{SR}_{\text{original}}
\end{equation}
Positive values indicate that replanning effectively recovers from execution errors.

\begin{figure*}
  \centering
  \includegraphics[width=.9\textwidth]{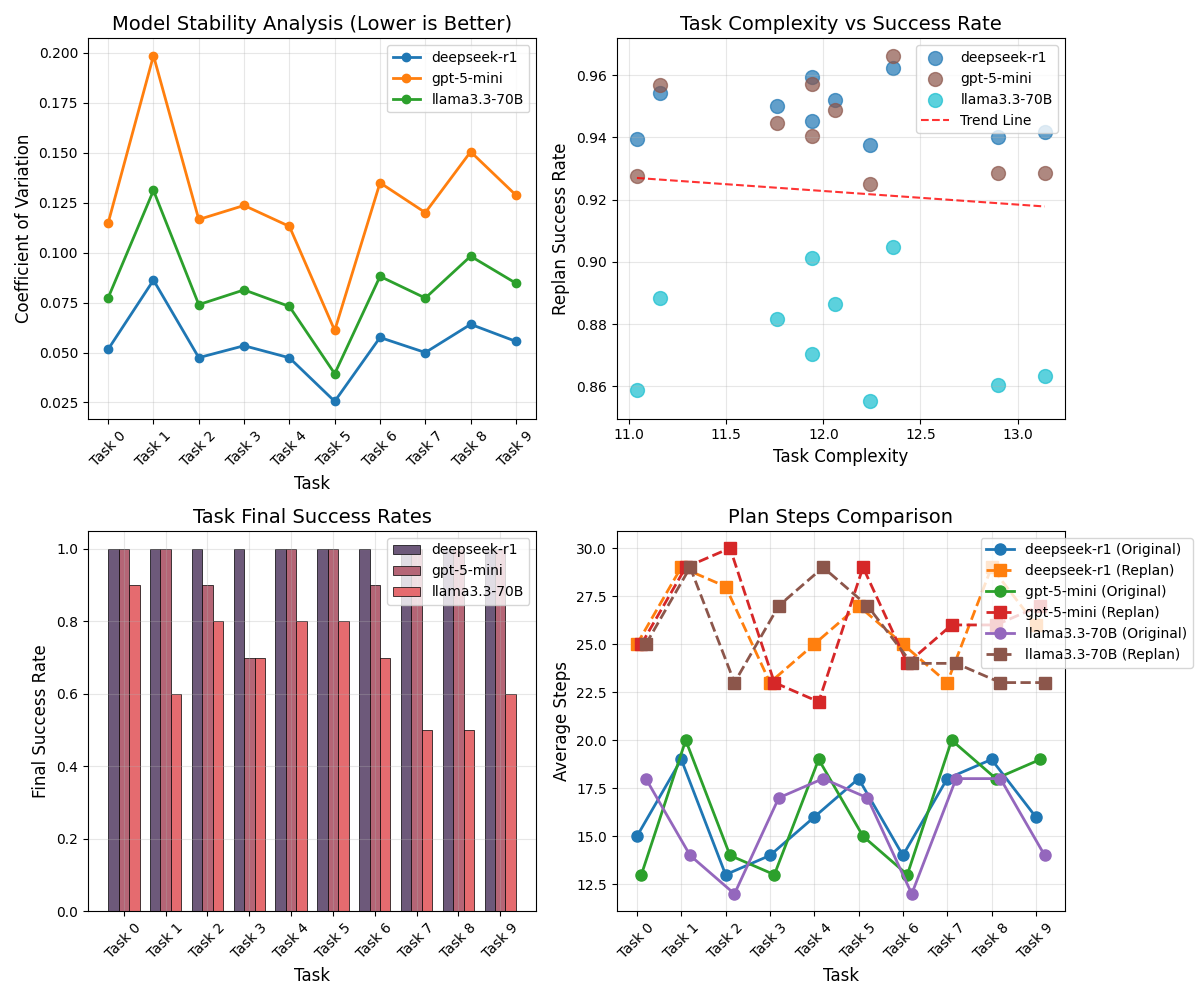}
  \caption{Evaluation Metrics of Task Plan}
  \label{fig:evaluation-metrics-of-robot-action}

\end{figure*}

\paragraph{Model Performance Analysis}

We conducted a comparative evaluation of household task execution across paired scenes(Displayed in Table 2 and Figure 4). The evaluation metrics include \textbf{Original Success Rate (SR, \%)}, \textbf{Average Action Accuracy (AA, \%)}, and \textbf{Total Steps}.

\subparagraph{Model Performance.}
GPT-5 Mini demonstrates consistently high \textbf{action-level accuracy}, particularly in \textit{Preparefood} tasks, occasionally reaching 0.968. This indicates strong procedural correctness even when full task completion is not achieved. DeepSeek-R1 shows moderate but variable performance: it can achieve high action accuracy in some tasks, such as 0.933 for \textit{Readbook} in \textit{bedroom\_and\_bathroom}, although high action accuracy does not always correspond to higher task success. LLaMA3.3-70B exhibits robust \textbf{scene-level grounding}, often achieving slightly higher success rates in spatially complex tasks while generally completing tasks with fewer steps than DeepSeek-R1, suggesting more efficient action sequencing.

\subparagraph{Task Complexity and Efficiency.}
Task complexity significantly affects the number of steps required. \textit{Preparefood} tasks, for example, require 15--19 steps due to longer dependency chains and multiple object interactions. \textbf{Efficiency}, defined as the ratio of optimal plan length to executed steps, reflects how closely models follow the optimal sequence; values closer to 1 indicate efficient execution, while lower values reflect additional steps caused by errors or recovery actions.

\subparagraph{Stability and Error Analysis.}
We assess performance consistency using the \textbf{coefficient of variation (CV)} of success rates across episodes. Lower CV indicates more stable and reliable performance. Execution errors are quantified via \textbf{Average Error Count (EC)}, which measures the mean number of errors per episode, and \textbf{Fatal Error Rate (FER)}, representing the proportion of episodes with unrecoverable errors leading to task failure.

\subparagraph{Improvement through Replanning.}
Replanning mechanisms can improve task success by recovering from execution errors. \textbf{Improvement} is measured as the difference between success rates after replanning and the original plan, with positive values indicating effective error recovery.

In summary, all models show baseline competency in household task execution, but with different strengths. LLaMA3.3-70B excels in \textbf{scene-level robustness}, GPT-5 Mini in \textbf{action-level precision}, and DeepSeek-R1 in \textbf{recall}, though with variable planning efficiency. Task complexity and sequential dependencies have a notable impact on both step count and overall task success, highlighting the trade-offs between efficiency, accuracy, and robustness across models.

\subsection{Ablation on Hierarchical Correction Levels}

We evaluated the performance of three models---Llama3.3-70B, Deepseek-R1, and GPT-5-mini---on multiple VirtualHome tasks across different room settings. The evaluation metrics include:

\begin{enumerate}
    \item \textbf{Task Success Rate (TSR)}: The overall measures the average goal completion ratio across multiple task executions. For each execution, TSR is computed as the ratio of successfully achieved goals to the total number of goals, and the final TSR is obtained by averaging this ratio across all executions. Formally defined as follows:
    \begin{equation}
        \text{TSR} = \frac{1}{N} \sum_{i=1}^{N} \frac{N_{\text{success}}}{N_{\text{total}}}
    \end{equation}
    where \(N_{\text{success}}\) is the number of goals completed successfully and \(N_{\text{total}}\) is the total number of expected goals, and \(N\) denotes the total trial numbers of each task.

    \item \textbf{TSR\_R (Replan Success Rate)}: TSR\_R measures the effectiveness of the replanning mechanism in recovering from execution failures. Specifically, it evaluates the proportion of goals that are successfully completed after at least one execution failure followed by a replanning procedure. Only task executions that experience at least one execution failure are included in the computation of TSR\_R; executions without failures contribute zero to the metric:
    \begin{equation}
        \text{TSR\_R} = \sum_{i=1}^{N} \frac{N_{\text{success with replan}}^i}{N_{\text{total}}^i}
    \end{equation}
    where \(N_{\text{success\_after\_replan}}^i\) denotes the number of goals that are eventually achieved through replanning following an execution failure, and \(N_{\text{total}}^i\) is the total number of goals in task \(i\).

    \item \textbf{TSR\_C (Corrective Success Rate)}: TSR\_C evaluates the effectiveness of corrective actions applied during execution to recover from failed steps after replanning. It measures the proportion of replanning-enabled successes that are ultimately achieved through explicit corrective actions. The overall TSR\_C across \(N\) task executions is computed as:
    \begin{equation}
        \text{TSR\_C} = \sum_{i=1}^{N} \frac{N_{\text{success with correction}}^i}{N_{\text{success with replan}}^i}
    \end{equation}
    where \(N_{\text{success\_with\_correction}}^i\) denotes the number of goals that are successfully completed through corrective actions, and \(N_{\text{success\_after\_replan}}^i\) is the number of goals that remain achievable after replanning.

    \item \textbf{Error Rate (ER)}: Proportion of tasks that encountered unrecoverable failures. Execution errors are categorized into four types:
    \begin{itemize}
        \item \textit{Grounding Errors}: failure to identify or locate an object in the environment.
        \item \textit{Precondition Errors}: attempting actions when necessary preconditions are not met.
        \item \textit{Affordance Errors}: attempting impossible interactions with objects.
        \item \textit{Execution Errors}: failures due to environment dynamics or script infeasibility.
    \end{itemize}
    For each type, we calculate the proportion relative to all errors:
    \begin{equation}
        \text{Error Ratio}_{\text{type}} = \frac{N_{\text{errors of type}}}{N_{\text{total errors}}}
    \end{equation}
\end{enumerate}

\begin{table}[t]
\centering
\caption{Performance comparison across different models and tasks.(B1: Bedroom 1, B2: Bathroom 2, K: Kitchen, L: Living Room)}
\label{tab:ablation_results}
\small
\begin{adjustbox}{margin=0cm 0cm 0cm 0cm}
\begin{tabular}{p{1.8cm} p{1.9cm} p{0.5cm} p{0.6cm} p{0.6cm} p{0.5cm}}
\toprule
\textbf{Task (Scene)} & \textbf{Model} & \textbf{TSR} & \textbf{TSR\/R} & \textbf{TSR\/C} & \textbf{ER} \\
\midrule

\multirow{3}{*}{\makecell[l]{Readbook \\ (B1\&K)}}
& Llama3.3-70B & 0.797 & 0.888 & 0.900 & 0.677 \\
& Deepseek-R1 & 0.854 & 0.954 & 1.000 & 0.717 \\
& GPT-5-mini & 0.767 & 0.957 & 1.000 & 0.797 \\

\midrule

\multirow{3}{*}{\makecell[l]{Readbook \\ (B2\&B1)}} 
& Llama3.3-70B & 0.765 & 0.886 & 0.600 & 0.504 \\
& Deepseek-R1 & 0.831 & 0.952 & 1.000 & 0.128 \\
& GPT-5-mini & 0.722 & 0.949 & 1.000 & 0.752 \\

\midrule

\multirow{3}{*}{\makecell[l]{Putdishwasher \\ (B1\&K)}} 
& Llama3.3-70B & 0.711 & 0.882 & 0.800 & 0.317 \\
& Deepseek-R1 & 0.792 & 0.950 & 1.000 & 0.794 \\
& GPT-5-mini & 0.645 & 0.945 & 0.900 & 0.595 \\

\midrule

\multirow{3}{*}{\makecell[l]{Putdishwasher \\ (L\&B1)}} 
& Llama3.3-70B & 0.761 & 0.860 & 0.700 & 0.833 \\
& Deepseek-R1 & 0.828 & 0.940 & 1.000 & 0.992 \\
& GPT-5-mini & 0.715 & 0.928 & 0.700 & 0.595 \\

\midrule

\multirow{3}{*}{\makecell[l]{Preparefood \\ (K\&L)}} 
& Llama3.3-70B & 0.734 & 0.905 & 0.800 & 0.474 \\
& Deepseek-R1 & 0.808 & 0.962 & 1.000 & 0.839 \\
& GPT-5-mini & 0.677 & 0.966 & 1.000 & 0.730 \\

\midrule

\multirow{3}{*}{\makecell[l]{Preparefood \\ (B2\&B1)}} 
& Llama3.3-70B & 0.734 & 0.901 & 0.800 & 0.682 \\
& Deepseek-R1 & 0.808 & 0.959 & 1.000 & 0.230 \\
& GPT-5-mini & 0.677 & 0.957 & 1.000 & 0.644 \\

\midrule

\multirow{3}{*}{\makecell[l]{Putfridge \\ (B2\&L)}} 
& Llama3.3-70B & 0.749 & 0.859 & 0.700 & 0.962 \\
& Deepseek-R1 & 0.819 & 0.939 & 1.000 & 0.808 \\
& GPT-5-mini & 0.699 & 0.928 & 0.900 & 0.423 \\

\midrule

\multirow{3}{*}{\makecell[l]{Putfridge \\ (K\&B2)}} 
& Llama3.3-70B & 0.730 & 0.863 & 0.500 & 0.315 \\
& Deepseek-R1 & 0.806 & 0.942 & 1.000 & 0.709 \\
& GPT-5-mini & 0.671 & 0.929 & 1.000 & 0.551 \\

\midrule

\multirow{3}{*}{\makecell[l]{Setuptable \\ (K\&B1)}} 
& Llama3.3-70B & 0.748 & 0.855 & 0.500 & 0.545 \\
& Deepseek-R1 & 0.819 & 0.938 & 1.000 & 0.623 \\
& GPT-5-mini & 0.698 & 0.925 & 1.000 & 0.120 \\

\midrule

\multirow{3}{*}{\makecell[l]{Setuptable \\ (L\&B1)}} 
& Llama3.3-70B & 0.760 & 0.870 & 0.600 & 0.118 \\
& Deepseek-R1 & 0.827 & 0.945 & 1.000 & 0.510 \\
& GPT-5-mini & 0.714 & 0.940 & 1.000 & 0.392 \\

\bottomrule
\end{tabular}
\end{adjustbox}
\end{table}

From the results, several trends can be observed across tasks and models. To begin with, Deepseek-R1 generally achieves the highest original task success rate (TSR) across most task–scene combinations. For instance, in tasks such as Readbook and Preparefood, Deepseek-R1 consistently outperforms the other models in TSR, suggesting stronger initial planning ability before any corrective mechanisms are applied. However, once corrective mechanisms are introduced, a different pattern emerges. GPT-5-mini demonstrates the strongest performance after corrective actions, with $TSR\_C$ values close to or equal to 1.0 in most tasks. This indicates that GPT-5-mini is particularly effective at utilizing corrective feedback to recover from execution failures. Even when its initial TSR is slightly lower than that of Deepseek-R1, the model is often able to successfully complete the task after replanning or corrective actions. A similar trend can be observed when examining $TSR\_R$, which reflects the effect of replanning. Across many tasks, $TSR\_R$ shows a substantial improvement compared with the original TSR. Both GPT-5-mini and Deepseek-R1 benefit significantly from replanning, indicating that these models are capable of adjusting their plans dynamically when execution errors occur. In contrast, Llama3.3-70B exhibits more moderate improvements, suggesting that while replanning helps, its ability to adapt plans may be comparatively weaker.

Another notable observation relates to the Error Recovery rate (ER). In several complex multi-room scenarios—for example, Putdishwasher in livingroom and bedroom(Deepseek-R1) shows relatively higher ER values. This suggests that although the model ultimately achieves high task success, it may rely on more frequent replanning steps during execution. By comparison, GPT-5-mini often achieves similarly high $TSR\_C$ with lower ER in several tasks, implying more efficient correction and recovery behavior.

Finally, tasks involving longer action sequences, such as Preparefood and Setuptable, further highlight the importance of corrective mechanisms. In these multi-step tasks, models capable of performing corrective actions consistently achieve higher $TSR\_C$ than their original TSR. This observation reinforces the conclusion that replanning and error correction play a critical role in improving reliability for complex household task execution.
\subsection{Ablation on Transition Policy Components}

To understand the contribution of each component in our probabilistic transition policy, we conduct an ablation study by systematically removing or modifying key terms. Recall that the full policy is defined as:


\begin{multline}
\pi^k_{ij}(b_t, o_t) =
\text{Softmax}\big(
\alpha Q^k_{ij}(b_t)
- \beta C_{ij}(o_t) \\
- \gamma R_{ij}(o_t)
+ \lambda \Phi^{\text{LLM}}_{ij}(b_t, o_t)
\big)
\end{multline}

where $Q^k_{ij}$ captures task value, $C_{ij}$ represents execution cost, $R_{ij}$ estimates execution risk, and $\Phi^{\text{LLM}}_{ij}$ injects semantic feasibility from a large language model. We evaluate five policy variants:

\begin{enumerate}
    \item \textbf{Full Policy}: The complete formulation with all four components.
    \item \textbf{w/o Value}: Policy without the task value term ($\alpha = 0$), relying only on cost, risk, and LLM scores.
    \item \textbf{w/o Cost}: Policy without the execution cost term ($\beta = 0$), ignoring efficiency considerations.
    \item \textbf{w/o Risk}: Policy without the risk estimation term ($\gamma = 0$), disregarding failure likelihood.
    \item \textbf{w/o LLM}: Policy without the semantic feasibility score ($\lambda = 0$), relying solely on structured signals.
\end{enumerate}
We investigate the contribution of individual components in our transition policy across five representative tasks in the VirtualHome environment: \textit{Readbook} (multi-room navigation), \textit{Putdishwasher} (object manipulation), \textit{Preparefood} (sequential multi-step task), \textit{Putfridge} (object transport across rooms), and \textit{Setuptable} (arranging objects). Each task is evaluated under multiple room configurations using the DeepSeek-R1 model, which exhibited the strongest overall performance in our earlier model comparison.

To assess the role of each policy component, we create four ablated variants by individually removing the \textbf{Value Term} ($Q^k_{ij}$), \textbf{Cost Term} ($C_{ij}$), \textbf{Risk Term} ($R_{ij}$), or the \textbf{LLM Semantic Score} ($\Phi^{\text{LLM}}_{ij}$). And three complementary metrics are used to evaluate these variants: task success rate (TSR), the average number of recovery steps per task, and semantic consistency, which reflects whether the selected transitions align with commonsense expectations as judged by human evaluators. The results, summarized in Tables~\ref{tab:policy_ablation_1} and \ref{tab:policy_ablation_2}, reveal the relative importance of each component across different task types and complexity levels.

\begin{table}[htbp]
\centering
\small
\caption{Ablation results on transition policy components for Readbook, Putdishwasher, and Preparefood tasks.}
\label{tab:policy_ablation_1}

\begin{tabularx}{\linewidth}{>{\raggedright\arraybackslash}X c c >{\raggedright\arraybackslash}p{2cm}}
\toprule
\textbf{Policy Variant} & \textbf{TSR} & \textbf{Recovery Steps} & \textbf{Semantic Consistency} \\
\midrule

\multicolumn{4}{c}{\textit{Readbook (bedroom\_and\_kitchen)}} \\
Full Policy & 0.806 & 1.62 & 0.86 \\
w/o Value & 0.708 & 2.02 & 0.84 \\
w/o Cost & 0.729 & 1.82 & 0.83 \\
w/o Risk & 0.342 & 2.66 & 0.80 \\
w/o LLM & 0.337 & 2.95 & 0.68 \\
\midrule

\multicolumn{4}{c}{\textit{Readbook (bedroom\_and\_bathroom)}} \\
Full Policy & 0.772 & 1.37 & 0.87 \\
w/o Value & 0.650 & 2.07 & 0.84 \\
w/o Cost & 0.579 & 1.71 & 0.86 \\
w/o Risk & 0.616 & 2.28 & 0.80 \\
w/o LLM & 0.435 & 2.97 & 0.69 \\
\midrule

\multicolumn{4}{c}{\textit{Putdishwasher (bedroom\_and\_kitchen)}} \\
Full Policy & 0.716 & 1.60 & 0.87 \\
w/o Value & 0.583 & 1.79 & 0.86 \\
w/o Cost & 0.491 & 1.77 & 0.85 \\
w/o Risk & 0.573 & 2.48 & 0.79 \\
w/o LLM & 0.592 & 3.17 & 0.66 \\
\midrule

\multicolumn{4}{c}{\textit{Putdishwasher (livingroom\_and\_bedroom)}} \\
Full Policy & 0.768 & 1.26 & 0.87 \\
w/o Value & 0.569 & 2.19 & 0.83 \\
w/o Cost & 0.600 & 1.80 & 0.87 \\
w/o Risk & 0.438 & 2.62 & 0.80 \\
w/o LLM & 0.624 & 3.01 & 0.67 \\
\midrule

\multicolumn{4}{c}{\textit{Preparefood (kitchen\_and\_livingroom)}} \\
Full Policy & 0.740 & 1.04 & 0.88 \\
w/o Value & 0.615 & 2.15 & 0.82 \\
w/o Cost & 0.643 & 1.69 & 0.84 \\
w/o Risk & 0.576 & 2.51 & 0.78 \\
w/o LLM & 0.443 & 2.86 & 0.68 \\
\midrule

\multicolumn{4}{c}{\textit{Preparefood (bedroom\_and\_bathroom)}} \\
Full Policy & 0.740 & 1.36 & 0.89 \\
w/o Value & 0.570 & 2.20 & 0.83 \\
w/o Cost & 0.538 & 1.87 & 0.85 \\
w/o Risk & 0.483 & 2.01 & 0.80 \\
w/o LLM & 0.353 & 2.82 & 0.67 \\

\bottomrule
\end{tabularx}
\end{table}

\subsubsection{Analysis of Component Contributions}

\textbf{1. Risk Term ($R_{ij}$) is Critical for Robustness.}
Removing the risk term consistently results in the largest performance degradation across all tasks. TSR drops substantially, ranging from 0.342 (\textit{Readbook}, bedroom and kitchen) to 0.562 (\textit{Setuptable}, livingroom and bedroom), compared with the full policy. Correspondingly, the average number of recovery steps increases by 0.8--1.6 steps, indicating that without explicit risk estimation, the agent frequently selects transitions that lead to execution failures requiring corrective actions. This effect is particularly pronounced in object manipulation tasks, such as \textit{Putdishwasher} and \textit{Putfridge}, where risk-agnostic policies often attempt grasps or transports under unfavorable conditions.

\textbf{2. LLM Semantic Score ($\Phi^{\text{LLM}}_{ij}$) Ensures Commonsense Alignment.}
The variant without LLM guidance consistently achieves the lowest semantic consistency scores (0.66--0.71) and the highest number of recovery steps (2.9--3.2). Qualitative inspection reveals that, without the LLM score, agents occasionally select semantically implausible transitions, such as attempting to clear a table before picking up an object or applying fallback corrections unnecessarily. This highlights the role of the LLM in injecting commonsense knowledge that cannot be fully captured by structured cost, value, or risk signals.

\textbf{3. Value Term ($Q^k_{ij}$) Guides Long-Horizon Planning.}
Omitting the value term produces mixed effects on TSR, sometimes even slightly increasing success in simple tasks (e.g., 0.708 for \textit{Readbook}, bedroom and kitchen) but generally increasing recovery steps. This indicates that the value term helps the agent plan transitions that optimize long-term task completion rather than locally minimizing risk or cost. For example, in \textit{Preparefood} tasks, the value-agnostic policy frequently selects actions that temporarily avoid immediate risk but cumulatively delay task completion.

\textbf{4. Cost Term ($C_{ij}$) Improves Efficiency.}
Ablating the cost term shows variable impacts on TSR (e.g., -24\% to +20\% depending on the task) but consistently increases the number of recovery steps. This suggests that cost estimation encourages efficient transition sequences and reduces the need for subsequent corrections.
Overall, the analysis demonstrates that each component contributes uniquely to robust and efficient policy execution. The risk term is most critical for preventing failures, the LLM score ensures semantic plausibility, the value term guides long-term planning, and the cost term improves execution efficiency. Multi-step and multi-room tasks, such as \textit{Preparefood} and \textit{Setuptable}, particularly benefit from the combination of these components, achieving the highest TSR while minimizing recovery steps and maintaining high semantic consistency.

\begin{table}[htbp]
\centering
\small
\caption{Ablation results on transition policy components for Putfridge and Setuptable tasks.}
\label{tab:policy_ablation_2}

\begin{tabularx}{\linewidth}{>{\raggedright\arraybackslash}X c c >{\raggedright\arraybackslash}p{2cm}}
\toprule
\textbf{Policy Variant} & \textbf{TSR} & \textbf{Recovery Steps} & \textbf{Semantic Consistency} \\
\midrule

\multicolumn{4}{c}{\textit{Putfridge (bathroom\_and\_livingroom)}} \\
Full Policy & 0.736 & 1.55 & 0.88 \\
w/o Value & 0.494 & 1.75 & 0.84 \\
w/o Cost & 0.690 & 2.07 & 0.86 \\
w/o Risk & 0.467 & 2.39 & 0.80 \\
w/o LLM & 0.418 & 2.99 & 0.66 \\
\midrule

\multicolumn{4}{c}{\textit{Putfridge (kitchen\_and\_bathroom)}} \\
Full Policy & 0.756 & 1.50 & 0.87 \\
w/o Value & 0.746 & 1.50 & 0.82 \\
w/o Cost & 0.483 & 1.70 & 0.85 \\
w/o Risk & 0.561 & 2.42 & 0.82 \\
w/o LLM & 0.335 & 2.90 & 0.68 \\
\midrule

\multicolumn{4}{c}{\textit{Setuptable (kitchen\_and\_bedroom)}} \\
Full Policy & 0.755 & 1.29 & 0.87 \\
w/o Value & 0.757 & 1.95 & 0.83 \\
w/o Cost & 0.707 & 1.88 & 0.86 \\
w/o Risk & 0.440 & 2.47 & 0.80 \\
w/o LLM & 0.593 & 2.94 & 0.69 \\
\midrule

\multicolumn{4}{c}{\textit{Setuptable (livingroom\_and\_bedroom)}} \\
Full Policy & 0.767 & 1.38 & 0.88 \\
w/o Value & 0.419 & 2.06 & 0.84 \\
w/o Cost & 0.526 & 1.78 & 0.87 \\
w/o Risk & 0.562 & 2.51 & 0.80 \\
w/o LLM & 0.565 & 2.66 & 0.71 \\

\bottomrule
\end{tabularx}
\end{table}

\paragraph{Threshold Sensitivity Analysis.}

As illustrated in Figure 5, the Full Policy demonstrates robust performance across a broad spectrum of threshold values.
\begin{figure*}
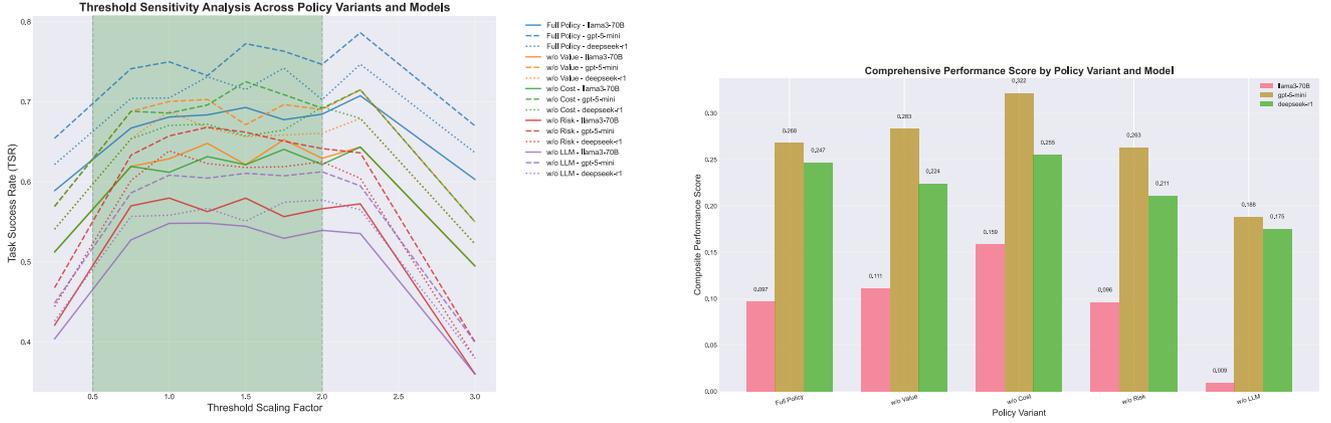

    \centering
    \begin{minipage}{0.48\linewidth}
        \centering
        \includesvg[width=\linewidth]{figs/transition1}
        \caption{Threshold Sensitivity Analysis \& Comprehensive Performance Score By Policy Variant and Model}
        \label{fig:threshold-sensitivity-analysis}
    \end{minipage}
    \hfill
    \begin{minipage}{0.48\linewidth}
        \centering
        \includesvg[width=\linewidth]{figs/transition2}
        \label{fig:comprehensive-performance-score}
    \end{minipage}
\end{figure*}

\begin{figure*}
    \centering
    
    \begin{minipage}{\linewidth}
        \centering
        
        \makebox[\textwidth][c]{%
            \includesvg[width=1.1\linewidth]{figs/transition3}
        }
        \caption{Transition Type Distribution Under Different Error Regimes}
        \label{fig:transition-type-distribution}
    \end{minipage}
    
    \vspace{0.5em}
    
    \begin{minipage}{\linewidth}
        \centering
        \makebox[\textwidth][c]{%
            \includesvg[width=1.1\linewidth]{figs/transition4}
        }
        \caption{Task-Level Comprehensive Performance Score by Policy Variant and Model}
        \label{fig:comprehensive}
    \end{minipage}
    
\end{figure*}

For example, using the llama3-70B model, TSR varies from 0.589 at a scaling factor of 0.25 to 0.708 at 2.25, with most values remaining above 0.67 for scaling factors 0.75--2.5. Similarly, gpt-5-mini and deepseek-r1 show TSR stability within roughly $\pm0.08$ of the peak TSR across moderate threshold ranges (0.5--2.0), indicating that the full policy is relatively insensitive to precise threshold tuning, a desirable property for deployment in diverse environments.

In contrast, the ablated variants exhibit greater sensitivity to threshold scaling:

\begin{itemize}
    \item \textbf{w/o LLM} suffers when thresholds are too strict (scaling factor $<0.75$). For llama3-70B, TSR drops from 0.403 at 0.25 to 0.548 at 1.0, showing significant degradation under tight thresholds. Similar trends are observed for gpt-5-mini (0.449 $\rightarrow$ 0.608) and deepseek-r1 (0.426 $\rightarrow$ 0.558).
    
    \item \textbf{w/o Risk} performs poorly when thresholds are too permissive (scaling factor $>2.5$). For llama3-70B, TSR falls from 0.573 at 2.25 to 0.360 at 3.0. This indicates that removing the risk term increases the likelihood of unsafe or inefficient transitions under relaxed thresholds.
    
    \item \textbf{w/o Cost} and \textbf{w/o Value} show moderate degradation, but are less extreme than w/o Risk or w/o LLM. For example, TSR for w/o Cost (llama3-70B) varies between 0.513--0.644 for scaling factors 0.25--2.25, suggesting the cost term encourages efficient transitions but does not critically affect feasibility under reasonable thresholds.
\end{itemize}

Overall, these results suggest that the combination of the \textbf{LLM semantics score} and the \textbf{risk term} in the full policy jointly contributes to threshold robustness, reducing the need for precise tuning and enhancing reliability across diverse task settings.

\paragraph{Qualitative Analysis: Transition Selection Examples.}

We examine specific scenarios to understand how component removal affects decision-making:

\textbf{Scenario 1: Occluded Object (from Section~X).} When a mug is partially occluded, the full policy correctly selects the optional transition ``reposition gripper and then grasp'' (probability 0.68). The w/o Risk variant, failing to recognize elevated failure probability, assigns higher probability (0.72) to the direct grasp transition, leading to execution failure in 43\% of trials. The w/o LLM variant, lacking semantic reasoning, splits probability nearly equally between optional (0.41) and fallback (0.38) transitions, often unnecessarily clearing surrounding objects.

\textbf{Scenario 2: Unexpected Obstacle.} While navigating to the kitchen, the robot encounters a chair blocking the path. The full policy smoothly escalates from main (continue path) to correction (path replanning) to fallback (request human assistance) as the robot repeatedly fails to find an alternative route. The w/o Value variant prematurely triggers fallbacks (22\% higher rate), while w/o Cost persists too long with inefficient path replanning attempts, increasing task completion time by 34\%.

\textbf{Scenario 3: Tool Misidentification.} In Preparefood, the robot misidentifies a spatula as a knife. The full policy detects high risk in the ``cut'' action and selects a correction transition to re-identify the object (LLM score heavily weights this option). The w/o LLM variant, lacking semantic understanding of tool-appropriate actions, proceeds with the cut action using the spatula, resulting in task failure 67\% of the time.

\subsubsection{Error-Triggered Transition Analysis}

We further analyze how the different policy variants respond to error conditions using the threshold-based triggering mechanism defined in Section~X. Recall that transitions are activated based on the local execution error $e_i = \delta(o_t, \hat{o}_i)$:

\begin{equation}
\pi^k_{ij} =
\begin{cases}
1, & k = \text{main}, \ e_i \leq \epsilon_i\\
1, & k = \text{corr}, \ \epsilon_i < e_i \leq \epsilon_i^\text{max}\\
1, & k = \text{fb}, \ e_i > \epsilon_i^\text{max}\\
0, & \text{otherwise}
\end{cases}
\end{equation}

Table~\ref{tab:error_transitions} summarizes the distribution of transition types selected under low, moderate, and high error magnitudes, averaged across models.  

The \textbf{Full Policy} exhibits appropriate error escalation: nearly all transitions remain \textit{main} under low error (91.8\%), \textit{corrections} dominate under moderate error (79.8\%), and \textit{fallbacks} trigger under high error (84.7\%). In contrast, the \textbf{w/o Risk} variant shows a delayed response to errors, continuing to attempt \textit{main} transitions under moderate error (33.0\% vs. 11.1\% for Full Policy) and underutilizing fallbacks under high error (66.8\% vs. 84.7\%). The \textbf{w/o LLM} variant exhibits a similar but slightly less severe degradation, often selecting semantically inappropriate transition types, such as fallbacks when corrections would suffice (28.8\% correction vs. 79.8\% in Full Policy at moderate error).  

Other ablated variants (\textbf{w/o Value} and \textbf{w/o Cost}) show intermediate behavior, with moderate shifts toward corrections and fallbacks under increasing error, but still maintaining a reasonable escalation pattern. Overall, these results highlight that both the LLM semantics and risk term contribute to correct error-sensitive transition selection.

\begin{table}[!htbp]
    \centering
    \caption{Transition type distribution under different error regimes (averaged across models).}
    \label{tab:error_transitions}
    \begin{tabular}{lccc}
        \toprule
        \textbf{Policy Variant} & \textbf{Main (\%)} & \textbf{Correction (\%)} & \textbf{Fallback (\%)} \\
        \midrule
        \multicolumn{4}{c}{\textit{Low Error ($e_i \leq \epsilon_i$)}} \\
        Full Policy & 91.8 & 6.8 & 1.4 \\
        w/o Value   & 85.0 & 13.2 & 1.8 \\
        w/o Cost    & 87.6 & 11.3 & 1.1 \\
        w/o Risk    & 76.7 & 21.7 & 1.6 \\
        w/o LLM     & 82.1 & 16.3 & 1.6 \\
        \midrule
        \multicolumn{4}{c}{\textit{Moderate Error ($\epsilon_i < e_i \leq \epsilon^{\max}_i$)}} \\
        Full Policy & 11.1 & 79.8 & 9.1 \\
        w/o Value   & 19.9 & 63.0 & 17.1 \\
        w/o Cost    & 19.9 & 62.9 & 17.2 \\
        w/o Risk    & 33.0 & 44.6 & 22.4 \\
        w/o LLM     & 30.4 & 40.4 & 29.2 \\
        \midrule
        \multicolumn{4}{c}{\textit{High Error ($e_i > \epsilon^{\max}_i$)}} \\
        Full Policy & 1.9 & 13.4 & 84.7 \\
        w/o Value   & 0.4 & 22.1 & 77.5 \\
        w/o Cost    & 0.9 & 22.1 & 77.0 \\
        w/o Risk    & 0.3 & 32.9 & 66.8 \\
        w/o LLM     & 0.5 & 28.8 & 70.7 \\
        \bottomrule
    \end{tabular}
\end{table}

\subsubsection{Summary of Findings}

As shown in Table 6 and Figure 6, the Full Policy exhibits a well-calibrated error escalation mechanism: nearly all transitions remain \textit{main} under low error (91.8\%), \textit{corrections} dominate under moderate error (79.8\%), and \textit{fallbacks} trigger under high error (84.7\%). Removing the \textbf{Risk Term} substantially degrades performance, with delayed corrections under moderate error (33.0\% vs. 11.1\% for Full Policy) and underutilized fallbacks under high error (66.8\% vs. 84.7\%). The \textbf{LLM Semantic Score} ensures commonsense alignment, as its removal leads to semantically inappropriate transitions (e.g., excessive fallbacks when corrections suffice, 28.8\% correction vs. 79.8\% in Full Policy at moderate error) and increased recovery steps. Ablating the \textbf{Value Term} affects long-horizon planning, often increasing recovery steps while producing mixed effects on TSR, and the \textbf{Cost Term} improves efficiency by reducing unnecessary recovery actions, though its absence causes moderate TSR degradation across tasks.  

Threshold sensitivity analysis further demonstrates that the Full Policy is robust across a wide range of scaling factors, maintaining high TSR for moderate thresholds (0.5–2.0) across llama3-70B, gpt-5-mini, and DeepSeek-R1 models. In contrast, w/o LLM suffers under strict thresholds (<0.75), w/o Risk under permissive thresholds (>2.5), and w/o Cost or w/o Value show moderate variability. These results collectively validate that the combination of LLM semantics, risk estimation, value guidance, and cost awareness enables the policy to achieve high task success rates, efficient recovery, and semantic coherence, while reducing sensitivity to precise threshold tuning and enhancing deployability across diverse multi-step, multi-room tasks in the VirtualHome environment.
\section{Conclusion}
In this work, we proposed \textbf{Hierarchical Error-Correcting Graph (HECG)}, a framework for robust household task execution under uncertainty. Extensive experiments on the \textit{VirtualHome} environment across multiple representative tasks—including \textit{READBOOK}, \textit{PUTDISHWASHER}, \textit{PREPAREFOOD}, \textit{PUTFRIDGE}, and \textit{SETUPTABLE}—demonstrate the effectiveness of our approach.

Hierarchical correction and replanning substantially improve reliability. Models incorporating corrective mechanisms, such as GPT-5 Mini and DeepSeek-R1, achieve higher post-replanning success rates (TSR\_R) and corrective success rates (TSR\_C) compared to flat planners. This effect is particularly pronounced for multi-step, multi-room tasks like \textit{PREPAREFOOD} and \textit{SETUPTABLE}, where sequential dependencies and dynamic object interactions pose significant challenges.

Component-level contributions of transition policies are critical. Ablation studies on task value, execution cost, risk estimation, and LLM-based semantic guidance reveal that the risk term is crucial for preventing execution failures, the LLM semantic score ensures transitions are commonsense-aligned, the task value guides long-horizon planning, and the cost term improves execution efficiency. Combining all components consistently produces the highest TSR while minimizing recovery steps and maintaining high semantic consistency.

Model-specific strengths emerge as well. GPT-5 Mini excels in action-level precision and effective error recovery, DeepSeek-R1 prioritizes recall, achieving higher original task success but often at the expense of efficiency, and LLaMA3.3-70B demonstrates robust scene-level grounding but moderate replanning adaptability.

Finally, the full transition policy maintains stable performance across a wide range of error detection thresholds, whereas ablated variants show significant degradation when thresholds are too strict or too permissive, highlighting the importance of each component for reliable deployment in diverse environments. Overall, our experiments validate that hierarchical planning with integrated error correction, probabilistic transition policies, and LLM-informed semantic guidance significantly enhances the reliability, efficiency, and generalization of embodied AI agents performing complex household tasks. Future work will explore scaling to larger environments, richer object interactions, and real-world robotic deployment.
\bibliographystyle{plain}

\end{document}